# Learning Tree-Structured Detection Cascades for Heterogeneous Networks of Embedded Devices

Hamid Dadkhahi
College of Information and Computer Sciences
University of Massachusetts Amherst
hdadkhahi@cs.umass.edu

Benjamin M. Marlin
College of Information and Computer Sciences
University of Massachusetts Amherst
marlin@cs.umass.edu

## ABSTRACT

In this paper, we present a new approach to learning cascaded classifiers for use in computing environments that involve networks of heterogeneous and resource-constrained, low-power embedded compute and sensing nodes. We present a generalization of the classical linear detection cascade to the case of tree-structured cascades where different branches of the tree execute on different physical compute nodes in the network. Different nodes have access to different features, as well as access to potentially different computation and energy resources. We concentrate on the problem of jointly learning the parameters for all of the classifiers in the cascade given a fixed cascade architecture and a known set of costs required to carry out the computation at each node. To accomplish the objective of joint learning of all detectors, we propose a novel approach to combining classifier outputs during training that better matches the hard cascade setting in which the learned system will be deployed. This work is motivated by research in the area of mobile health where energy efficient real time detectors integrating information from multiple wireless on-body sensors and a smart phone are needed for real-time monitoring and the delivery of just-in-time adaptive interventions. We evaluate our framework on mobile sensor-based human activity recognition and mobile health detector learning problems.

## 1 INTRODUCTION

The field of mobile health or mHealth [9] aims to leverage recent advances in wearable on-body sensing technology and mobile computing to develop systems that can monitor health states and deliver just-in-time adaptive interventions [11]. These systems involve networks of heterogeneous on-body sensing devices that typically communicate wirelessly with a smart phone. Each device in the system typically has access to different sensor data streams and has different computational capabilities and energy resources.

mHealth research currently targets a wide range of health problems including stress [13], smoking [1, 17], overeating [19], and even drug use [7, 12]. These applications often use one or more wearable sensing devices including smart watches like the Samsung Gear or Pebble watch, and chest band sensors like the Zephyr BioHarness. These embedded devices have limited energy and compute resources due to their small form factors. The wearable sensors are linked with a smart phone (typically using Bluetooth) that has greater, but still limited, energy and compute resources. Time and energy costs are also incurred when transmitting data to the smart phone for aggregation.

However, current data analytics research in mHealth focuses almost exclusively on passive data collection followed by offline data analysis based on common machine learning models and algorithms including support vector machines [5] and random forests [3]. This research on detection models implicitly assumes that features from all sensors are available simultaneously, that compute resources are unbounded, and that results do not need to be delivered in real time. While this research is an important first step in establishing detector performance in the absence of real-world constraints, the development of practical systems that can support real-time monitoring and just-in-time-adaptive interventions while operating for long time periods in energy constrained computing environments requires an approach to data analytics that respects the inherent resource constraints of mHealth systems.

In this paper, we address the foundational problem of providing detector learning approaches for the mHealth domain that directly address the challenges imposed by data locality, energy limitations, and computational constraints. In particular, we propose a novel approach to learning cascaded classifiers for use in computing environments that involve networks of heterogeneous and resource-constrained, low-power embedded compute and sensing nodes. We present a generalization of the classical linear detection cascade to the case of tree-structured cascades where different branches of the tree execute on different physical compute nodes in the network. We model the fact that different nodes have access to different features, as well as access to potentially different computation and energy resources. We concentrate on the problem of jointly learning the parameters for all of the classifiers in the cascade given a fixed cascade architecture and a known set of costs required to carry out the computation at each stage in the cascade.

To accomplish the objective of joint learning of all detectors in a tree-structured cascade, we propose a novel approach to combining classifier outputs during training. Our approach can be seen as a significant generalization of the soft cascade learning framework [14] to the case of tree-structured cascades. We simultaneously modify the classifier combination and regularization functions to better match the hard cascade setting in which the learned system will be deployed. We refer to our general cascade learning approach, which also applies to classical linear cascades, as the *Firm Cascade Framework* to emphasize its goal of better modeling the hard decisions that occur when models are deployed.

We present experiments comparing our firm cascade framework to the soft cascade framework as well as to single-stage models using data from the smoking detection domain. This data set includes sensor data streams from both a wrist-worn actigraphy sensor and a respiration chest band sensor. We further investigate the performance of the proposed firm cascade framework on two activity recognition datasets. We explore a variety of cascade architectures including two and three stage linear cascades and tree-structured cascades. Our results show that tree-structured cascades with independent computation in different branches can be used in place of linear cascades in this domain with little loss of accuracy or computational efficiency. Our results also show that the firm cascade learning framework outperforms the soft cascade framework either in terms of accuracy or cost across a wide range of settings when used to train the same cascade architecture.

## 2 RELATED WORK

A classical linear classifier cascade is a collection of models that are applied in sequence to classify a data instance. In order for a data instance to be classified as positive, it must be classified as positive by all stages in the cascade. If any stage in the cascade rejects a data instance, processing of that instance immediately stops and it is classified as a negative instance. For highly class-imbalanced data, cascades can lead to substantial computational speedups.

Perhaps the most well-known work on classifier cascade learning is the Viola-Jones face detection framework [21]. This framework trains a classification model for each stage sequentially using a boosting algorithm [6]. Each stage is trained by boosting single-feature threshold classifiers by training only on the positive examples propagated by the previous stage. The bias of the final boosted model for each stage is then adjusted to minimize the number of false negatives. The Viola-Jones cascade can achieve real-time face detection by quickly rejecting the vast majority of sub-windows in an image that do not contain a face.

Subsequent work on boosting-based learning for cascades has focused on a number of shortcomings of the Viola-Jones cascade including extensions of adaboost for improved design of the cascade stages, joint training instead of greedy stage-wise training, and methods for learning optimal configurations of a boosted cascade including the number of boosting rounds per stage and the number of total stages. Saberian et al. present an excellent discussion of this work [16].

An alternative to boosting for cascade learning is the noisy-AND approach [10]. In this framework, the probability that an instance is classified as positive is given by the product of the output probabilities of an ensemble of probabilistic base classifiers (often logistic regression models). If any element of the ensemble predicts a negative label for a data instance, the instance will receive a negative label. The models in the ensemble are trained jointly using the cross-entropy loss applied to the product of their probabilities. For deployment as a cascade, the learned models must be placed in sequence in some way. A disadvantage of the noisy-AND approach is that there is no explicit penalization related to how many stages a data case propagates through before it is rejected as a negative example.

Raykar et al. proposed a modification to the noisy-AND approach that retains the cross-entropy/noisy-AND objective, but adds a penalty term to penalize the joint model based on the number of stages required to reject an example [14]. They refer to their approach as the "soft cascade." The primary disadvantage of their approach is that the cascade is still operated using hard decisions once deployed. This is not well-matched to the training objective, which retains the noisy-AND classifier combination rule. Our firm cascade framework significantly generalizes the soft cascade framework of Raykar et al. to the case of tree-structured cascades. Our framework also simultaneously modifies the classifier combination and regularization functions to better match the hard cascade setting in which the learned system will be deployed.

Our proposed framework is also similar to recent work on formulations of cost-sensitive classification that seek to trade off feature extraction cost against accuracy. More specifically, [20] and [22] consider a directed acyclic graph with feature/sensor subsets as nodes. Each node chooses whether to acquire features from a given sensor or to classify an instance using the available measurements. This problem is formulated as an empirical risk minimization problem for which an efficient algorithm based on dynamic programming is proposed in [22]. [25] offers an extension of stage-wise regression to the feature extraction cost minimization problem and is applicable to either regression or multi-class classification. The Cost-Sensitive Cascade of Classifiers (CSCC) and Cost-Sensitive Tree of Classifiers (CSTC) frameworks [4, 23, 24] consider a similar setting with a set of linear classifiers used at nodes in a chain or tree with the goal of minimizing feature extraction costs while preserving accuracy. Our framework differs from these in that the availability of features at given stages in our setting is dictated by external constraints imposed by the network topology of sensors and devices. We also focus on the trade-off between the total computational cost of classifying instances and classification accuracy, not just the cost of feature extraction/acquisition.

## 3 THE FIRM CASCADE FRAMEWORK

In this section, we first develop the firm cascade framework for the classical case of a binary linear cascade. We then generalize the framework to the case of tree-structured binary cascades.

### 3.1 Linear Cascade Architecture

To begin, assume we wish to learn a soft linear cascade model consisting of $L$ stages. We define a probabilistic classifier $P_l(y|\mathbf{x})$ for each stage $l$ where $\mathbf{x} \in \mathbb{R}^D$ is the feature vector, and $y \in \{0, 1\}$ is the binary class label. We let the probabilistic output of the cascade be $P_*(y|\mathbf{x})$. In the noisy-AND and soft cascade frameworks described in the previous section, $P_*(y|\mathbf{x})$ is defined as shown below:

$$P_*(y|\mathbf{x}) = \prod_{l=1}^{L} P_l(y|\mathbf{x}) \tag{1}$$

Our proposed firm cascade framework is based on an alternative combination rule that better reflects the idea that when a probabilistic cascade is operated in hard decision mode at deployment time, the output of each stage of the cascade gates the computation of the subsequent stage. In particular, for $1 \leq l \leq L-1$, a data case $\mathbf{x}$ is only passed from stage $l$ to stage $l+1$ of the cascade if $P_l(y|\mathbf{x}) > 0.5$, otherwise the data case is predicted to belong to class 0 and the computation halts at stage $l$. Our combination rule for a general linear cascade is given below. We use the shorthand

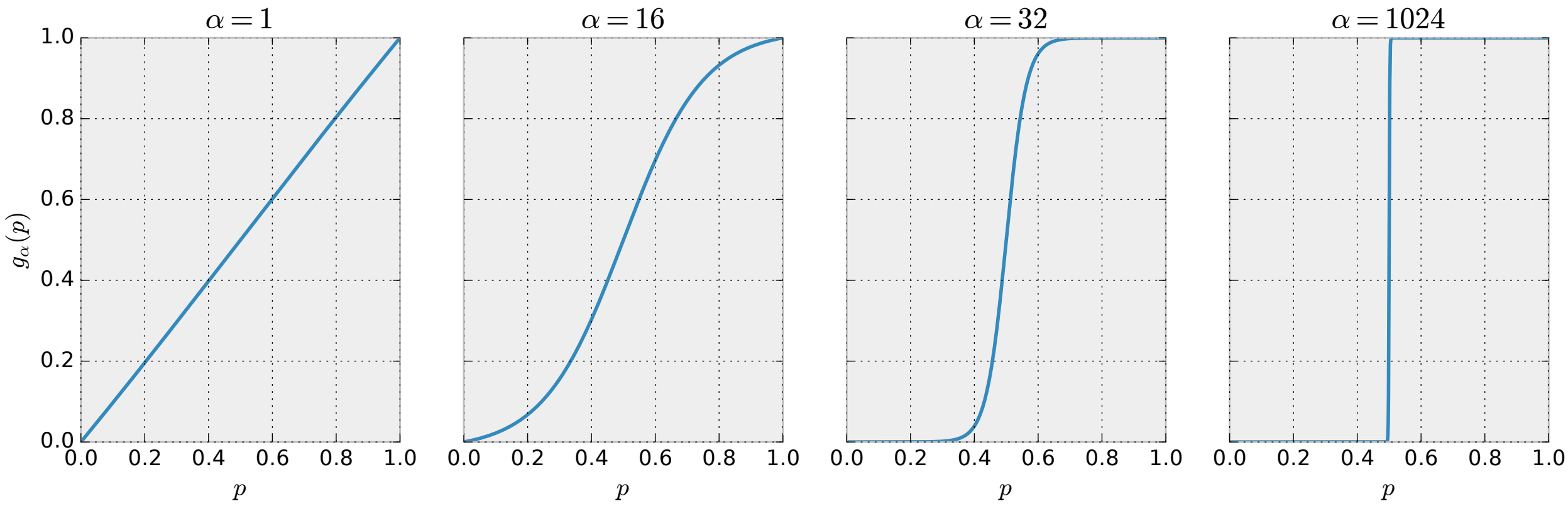

**Figure 1: Examples of the gating function $g_\alpha(p)$ at different values of the parameter $\alpha$.**

$p_l = P_l(y|\mathbf{x})$ to simplify the notation.

$$P_*(y|\mathbf{x}) = \sum_{l=1}^{L} \theta_l \cdot p_l \tag{2}$$

$$\theta_l = \begin{cases} (1 - g_\alpha(p_l)) \prod_{k=1}^{l-1} g_\alpha(p_k) & l < L \\ \prod_{k=1}^{L-1} g_\alpha(p_k) & l = L \end{cases} \tag{3}$$

$$f_\alpha(p) = \frac{1}{1 + \exp(-\alpha(p - 0.5))} \tag{4}$$

$$g_\alpha(p) = \frac{f_\alpha(p) - f_\alpha(0)}{f_\alpha(1) - f_\alpha(0)} \tag{5}$$

Equations 2 to 5 show that our proposed model takes the form of a mixture of experts [8] with highly specialized mixture weights. The effect of these mixture weights is to place nearly all of the weight in the mixture either on the output of the first stage in the cascade that classifies an instance as class 0, or on the output of the classifier in the last stage of the cascade. This is accomplished using the function $g_\alpha(p)$ shown in Equation 5 with a moderately large value of $\alpha$.

The function $g_\alpha(p)$ applies a normalized logistic nonlinearity to the input probability $p$ to approximate the hard decision function $\mathbb{I}_{0.5}(p)$ that is used at each stage of a linear cascade at deployment time ($\mathbb{I}_{0.5}(p) = 1$ if $p > 0.5$ and is 0 otherwise).[1] We show examples of the function $g_\alpha(p)$ for different values of $\alpha$ in Figure 1. We can see that the function satisfies $g_\alpha(0) = 0$ and $g_\alpha(1) = 1$ for all $\alpha$ due to the normalization term. As $\alpha$ increases, $g_\alpha(p)$ provides an increasingly accurate approximation to the hard step function $\mathbb{I}_{0.5}(p)$ while remaining smooth and differentiable. Importantly, this approximation is amenable for use with standard continuous optimization methods. In practice we use $\alpha = 32$ in our experiments, but we observe broad insensitivity to the choice of this parameter (see Figure 4 and related experiments for details).

[1] We note that the normalization of $f_\alpha(p)$ only impacts the gating function at small values of $\alpha$ ($\alpha < 8$), and is used to enforce $f_\alpha(0) = 0$ and $f_\alpha(1) = 1$.

In Equation 6, we give an example of the explicit form of a three-stage linear cascade to further clarify the cascade design.

$$P_*(y|\mathbf{x}) = (1 - g_\alpha(p_1)) \cdot p_1 + g_\alpha(p_1)(1 - g_\alpha(p_2)) \cdot p_2 + g_\alpha(p_1) g_\alpha(p_2) \cdot p_3 \tag{6}$$

If the output of the first stage $p_1$ is less than 0.5, $g_\alpha(p_1)$ will be close to zero and the output of the cascade will be $P_*(y|\mathbf{x}) \approx p_1$. If the output of the first stage is greater than 0.5, but the output of the second stage is less than 0.5, then $g_\alpha(p_1)$ will be close to 1 while $g_\alpha(p_2)$ will be close to 0 and the output of the cascade will be $P_*(y|\mathbf{x}) \approx p_2$. Finally, if both $p_1$ and $p_2$ are greater than 0.5, then both $g_\alpha(p_1)$ and $g_\alpha(p_2)$ will be close to 1 and the output of the cascade will be $P_*(y|\mathbf{x}) \approx p_3$. Thus, the probability output by the cascade will be approximately equal to either the output of the first stage $l$ to reject a data instance with $p_l < 0.5$, or the output of the final stage, $p_L$. As noted previously, this model can be viewed as a self-gated mixture of experts since the usual independent gating function is replaced by a gating function based on the outputs of the experts themselves.

Unlike the majority of work on classifier cascade learning that assumes the same base classifier is applied at all stages using different features, we consider architectures where different stages can use different base classifiers with different computational requirements as well as different features with the idea that these classifiers will run on different physical devices with different computational and energy resources as well as access to different sensor data streams. When each stage in the cascade is either a logistic regression classifier or a feedforward neural network (multi-layer perceptron) with a logistic output, the complete firm cascade model can also be viewed as a single multi-layer neural network model with a specialized output non-linearity that performs a soft selection among the outputs of the models from the $L$ stages.

## 3.2 The Tree-Structured Cascade Architecture

In this section, we generalize the linear cascade architecture described previously to the case of directed tree-structured cascades. The motivation for considering this extension is a deployment setting involving multiple heterogeneous sensing and computation devices all potentially operating in parallel. Each device runs its own linear cascade. If the cascade on a given device has positive output, then that device forwards its output and any needed features to the next device node in the directed tree (or produces a final output). In the mHealth setting, the network typically consists of a collection of wearable sensors that communicate only with a smart phone. The underlying device network thus has a star topology. In this section, we focus on this particular device network architecture for concreteness, but the same components that we introduce could be used to design cascades with more complex tree structures.

To begin, we assume we have access to a total of $D + 1$ devices $1, \ldots, D+1$ with device $D+1$ corresponding to the smart phone. Each device $d$ runs a linear cascade with $L_d$ stages. We let the probability computed by the classifier at stage $l$ on device $d$ be $P_l^d(y|\mathbf{x})$. We denote the output of the cascade for device $d$ by $P_*^d(y|\mathbf{x}) = p_*^d$. For devices $1 \leq d \leq D$, $P_*^d(y|\mathbf{x})$ is defined as shown in Equation 2. In hard decision mode, we assert logical-AND semantics when combining the final outputs from devices $1 \leq d \leq D$. This requires that all of the devices $1 \leq d \leq D$ predict that a data instance is positive in order for further processing to occur on device $D + 1$. During learning, we approximate the gating required by the logical-AND sematics using the combination function $\prod_{d=1}^{D} g_\alpha(p_*^d)$ with $g_\alpha(p)$ defined as in Equation 5. If any of the probabilities $p_*^d$ is less than 0.5, the combination will be much closer to 0 than a product of the raw probabilities $p_*^d$. Below, we define the final output probability $P_*(y|\mathbf{x})$ for a basic tree-structured cascade based on this combination rule and the application of a further linear cascade operating on device $D + 1$.

$$P_*(y|\mathbf{x}) = \sum_{l=1}^{L_{D+1}} \theta_l \cdot p_l^{D+1} \tag{7}$$

$$\theta_l = \begin{cases} (1 - \prod_{d=1}^{D} g_\alpha(p_*^d)) & l = 1 \\ \left(1 - g_\alpha(p_l^{D+1})\right) \prod_{k=1}^{l-1} g_\alpha(p_k^{D+1}) \prod_{d=1}^{D} g_\alpha(p_*^d) & 2 \leq l < L^{D+1} \\ \prod_{k=1}^{l-1} g_\alpha(p_k^{D+1}) \prod_{d=1}^{D} g_\alpha(p_*^d) & l = L^{D+1} \end{cases} \tag{8}$$

We note that the form of the final output probability $P_*(y|\mathbf{x})$ is similar to the linear cascade introduced in the previous section except for the effect of its first stage, which accomplishes the combination of the outputs from the previous $D$ devices. As defined in Equation 8, $\theta_1$ will be close to 1 if any of the first $D$ devices rejects the data instance. The probabilistic output of stage 1 of the cascade on device $D + 1$ is then defined to be the noisy-AND of the probabilities of the first $D$ devices: $p_1^{D+1} = P_1^{D+1}(y|\mathbf{x}) = \prod_{d=1}^{D} p_*^d$. The probability $p_l^{D+1} = P_l^{D+1}(y|\mathbf{x})$ for stages $2 \leq l \leq L_{D+1}$ on device $D + 1$ is defined by the local model in that stage, as for our earlier linear cascade model. The mixture weights $\theta_l$ for the later stages are also similar to the linear case, as seen in Equation 8, but include a $\prod_{d=1}^{D} g_\alpha(p_*^d)$ term that models the fact that the outputs of all of the first $D$ devices need to be positive for further stages of the cascade on device $D + 1$ to run.

## 3.3 Learning Cascade Models

To learn the linear firm cascade model, we maximize the log likelihood of the cascade output $P_*(y|\mathbf{x})$ as defined in Equation 2 (equivalent to minimizing the cross entropy loss), subject to a per-instance regularizer $r(y_n, \mathbf{x}_n)$. The objective function is shown below where the data set is $\mathcal{D} = \{(y_n, \mathbf{x}_n) | 1 \leq n \leq N\}$ and $N$ is the number of data instances.

$$\mathcal{L}(\mathcal{D}) = \sum_{n=1}^{N} \ell(y_n, \mathbf{x}_n) - \lambda r(y_n, \mathbf{x}_n) \tag{9}$$

$$\ell(y, \mathbf{x}) = y \log P_*(y|\mathbf{x}) + (1 - y) \log(1 - P_*(y|\mathbf{x})) \tag{10}$$

$$r(y, \mathbf{x}) = \kappa_1 + \sum_{l=2}^{L} \kappa_l \prod_{k=1}^{(l-1)} g_\alpha(P_k(y|\mathbf{x})) \tag{11}$$

Again, with a large value of $\alpha$, $g_\alpha(P_l(y|\mathbf{x}))$ will be approximately 0 for stages that output values that are less than 0.5, and will be approximately 1 for stages that are greater than 0.5. Thus, this regularizer applies a penalty approximately equal to the total cost of executing the number of stages actually used in the cascade to classify a given instance, where $\kappa_l$ is the cost per stage. It is similar to the penalty function used in [14], but is a better match to a hard cascade due to approximating the step function with the $g_\alpha()$ function.

To learn the tree-structured firm cascade model, we maximize the log likelihood of the final tree-structured cascade output $P_*(y|\mathbf{x})$ as defined in Equation 7. We again apply a per-instance regularizer $r(y_n, \mathbf{x}_n)$, which now has a more complex form due to the fact that we must take into account the cost of running multiple cascades on different devices in parallel. We let $\kappa_l^d$ be the cost of running stage $l$ of the classifier cascade for device $d$. The objective function is shown below where the data set is again $\mathcal{D} = \{(y_n, \mathbf{x}_n) | 1 \leq n \leq N\}$, $N$ is the number of data instances, $D + 1$ is the number of devices, and $L_d$ is the number of stages per device $d$.

$$\mathcal{L}(\mathcal{D}) = \sum_{n=1}^{N} \ell(y_n, \mathbf{x}_n) - \lambda r(y_n, \mathbf{x}_n) \tag{12}$$

$$r(y, \mathbf{x}) = \sum_{d=1}^{D} \left( \kappa_1^d + \sum_{l=2}^{L_d} \kappa_l^d \prod_{k=1}^{(l-1)} g_\alpha(P_k^d(y|\mathbf{x})) \right) + \kappa_1^{D+1} + \sum_{l=2}^{L_{D+1}} \kappa_l^{D+1} \prod_{d=1}^{D} g_\alpha(P_*^d(y|\mathbf{x})) \prod_{k=2}^{(l-1)} g_\alpha(P_k^{D+1}(y|\mathbf{x})) \tag{13}$$

Unlike most earlier work on boosted cascades, there is a direct mapping between the features available at a given stage and the hardware that stage runs on, so there is much more limited flexibility in the assignment of features to stages. The computational resources on a given device may also dictate the complexity of the classification models that can be run on that device. As a result, we focus on the problem of jointly optimizing the parameters of fixed cascade architectures as opposed to automatically optimizing the cascade architecture itself (often referred to as the cascade design problem). In our experiments, we use either logistic regression models or neural network models at each stage in each cascade. We implement the framework in Theano [18], which allows for rapid specification and testing of different cascade architectures.

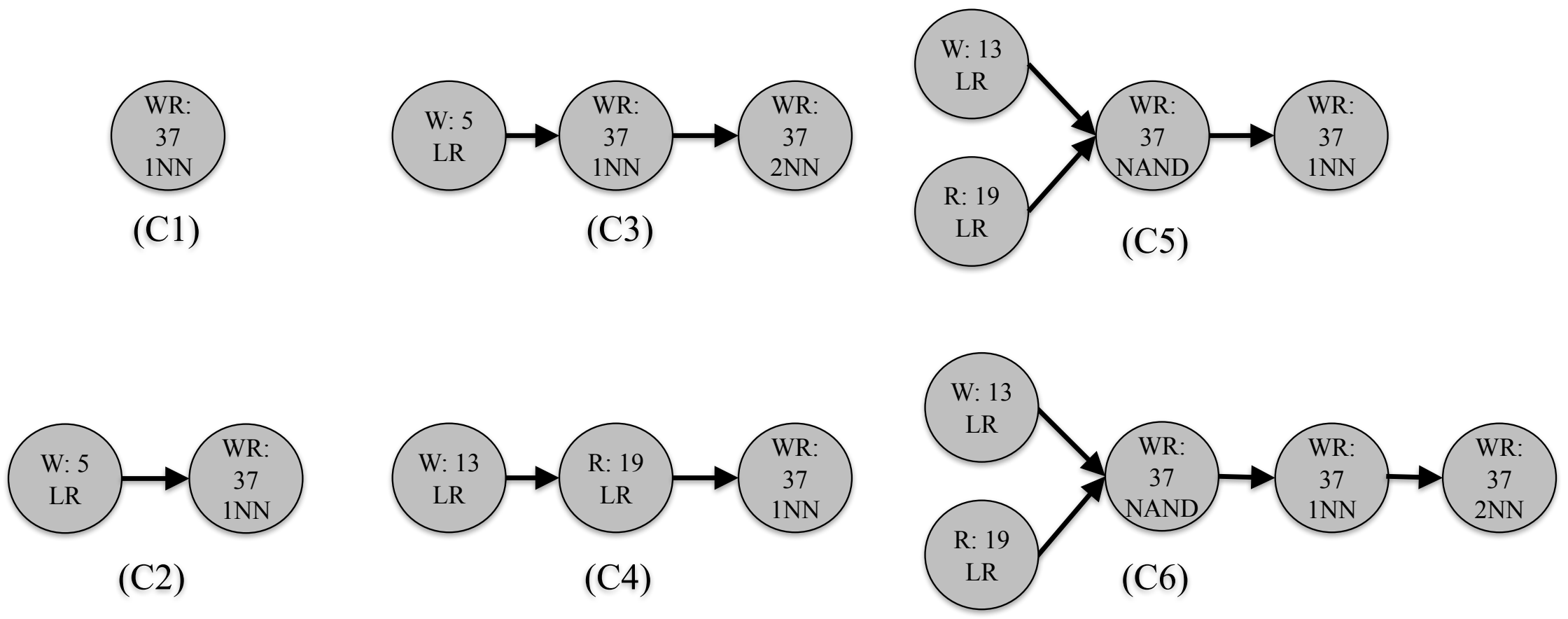

**Figure 2: This figure shows the six cascade architectures used in the empirical evaluation. Each node corresponds to a classifier in the cascade and is annotated with the type features used (W for wristband, R for respiration sensor, WR for both), the number of features used, and the type of classifier used (LR for logistic regression, 1NN for a one hidden layer classifier, 2NN for a two hidden layer classifier).**

We use RMSProp to learn the model parameters by maximizing the appropriate (linear or tree-structured) objective function $\mathcal{L}(\mathcal{D})$.

Finally, we note that while the complete set of models used in a given cascade can be optimized jointly using the objective function described above, we can also exploit the fact that the models used in later stages of the cascade are often increasingly powerful to develop a reverse stage-wise initialization. Specifically, for each device $d$, we initialize training by learning the models in reverse order from stage $L_d$ to stage 1, with the model for stage $l$ being able to depend on the downstream performance of stages $l+1$ to $L_d$ as well as models from device $D+1$. We use this initialization combined with fine tuning the cascade using joint training in the experiments that follow.

## 4 EXPERIMENTS AND RESULTS

In this section, we present experimental results comparing our proposed firm cascade architecture to the soft cascade of [14]. We consider the problem of smoking puff detection from wearable sensor data [17], and two different human activity recognition problems based on wearable sensor data [2, 15].

### 4.1 Smoking Puff Detection

As a test bed, we first use the PuffMarker smoking puff detection dataset from [17], which is highly class imbalanced. In this domain, simple feature extraction and detection could run on the wearable sensors, while more complex feature extraction and detection functions must run on a smart phone. In the PuffMarker[2] data set, each data case consists of 37 features. 19 features are computed from a respiratory inductance plethysmography sensor data stream, and 13 features are computed from accelerometer and gyroscope sensors on a wrist band. The remaining 5 features are computed from combinations of both wrist and respiration data. Overall, there are 3836 data cases in the PuffMarker dataset.

Figure 2 shows a graphical representation of the example cascade architectures that we consider in the experiments. We compare a single-stage model $C1$ to several linear and tree-structured cascades $C2-C6$. We train cascades $C2$ to $C6$ using both our firm cascade learning approach and the soft cascade approach. For cascades $C5$, and $C6$, when training using the soft cascade framework (which did not consider the case of tree-structured cascades) we apply the noisy-AND function over all nodes to obtain $P_*(y|\mathbf{x})$ and learn using an alternate version of our tree-based regularizer that uses the raw per-stage probabilities. This regularizer generalizes the original soft cascade regularizer to the case of trees without applying the gating function used in our firm cascade framework.

For the single-stage baseline model, we use a one-hidden-layer neural network (1LNN) with $K = 10$ hidden units and all 37 features. In all cascade models, we use logistic regression (LR) in the first stage. For cascade models $C2$ and $C3$, in the first stage, we consider 5 features obtained via the basis expansion $\Phi : [x, y] \rightarrow [x, y, x^2, y^2, xy]$ applied to roll ($x$) and pitch ($y$) features computed from the accelerometer data streams. This feature set is suggested by results in [17]. For $C2$, we use a one-hidden-layer neural network (1LNN) with $K = 10$ hidden units in the second stage. For the three-stage model $C3$, we use a one-layer neural network (1LNN) with $K_1 = 3$ hidden units as the second-stage classifier and a two-layer neural network (2LNN) with $K_1 = 10$ and $K_2 = 20$ hidden units as the third-stage classifier. All models in the second and third stages use logistic non-linearities and all 37 features. For the cascade model $C4$, we use LR in first and second stages, where we use 13 wrist and 19 respiration features, respectively. In the third stage, we use a 1LNN with $K = 10$. For the tree cascade model $C5$, we use LR in both branches of the tree, and use the 13 wrist features in one branch, and the 19 respiration features in the other. In the final stage, we use a 1LNN with $K = 10$. For the tree cascade model $C6$, we use LR with 13 wrist features in the wrist branch and LR with 19 respiration features in the respiration branch. We use a 1NN with $k_1 = 3$ followed by a 2LNN with $K_1 = 10$ and $K_2 = 20$ on the final device. Preliminary testing was used to identify the hidden layer sizes. Using larger hidden layer sizes tends to either result in lower accuracy due to over-fitting or similar accuracy, but increased time. We assume a cost-per stage that is proportional to the compute time for each stage.

[2]Note that we used the dataset exactly as explained in the PuffMarker paper.

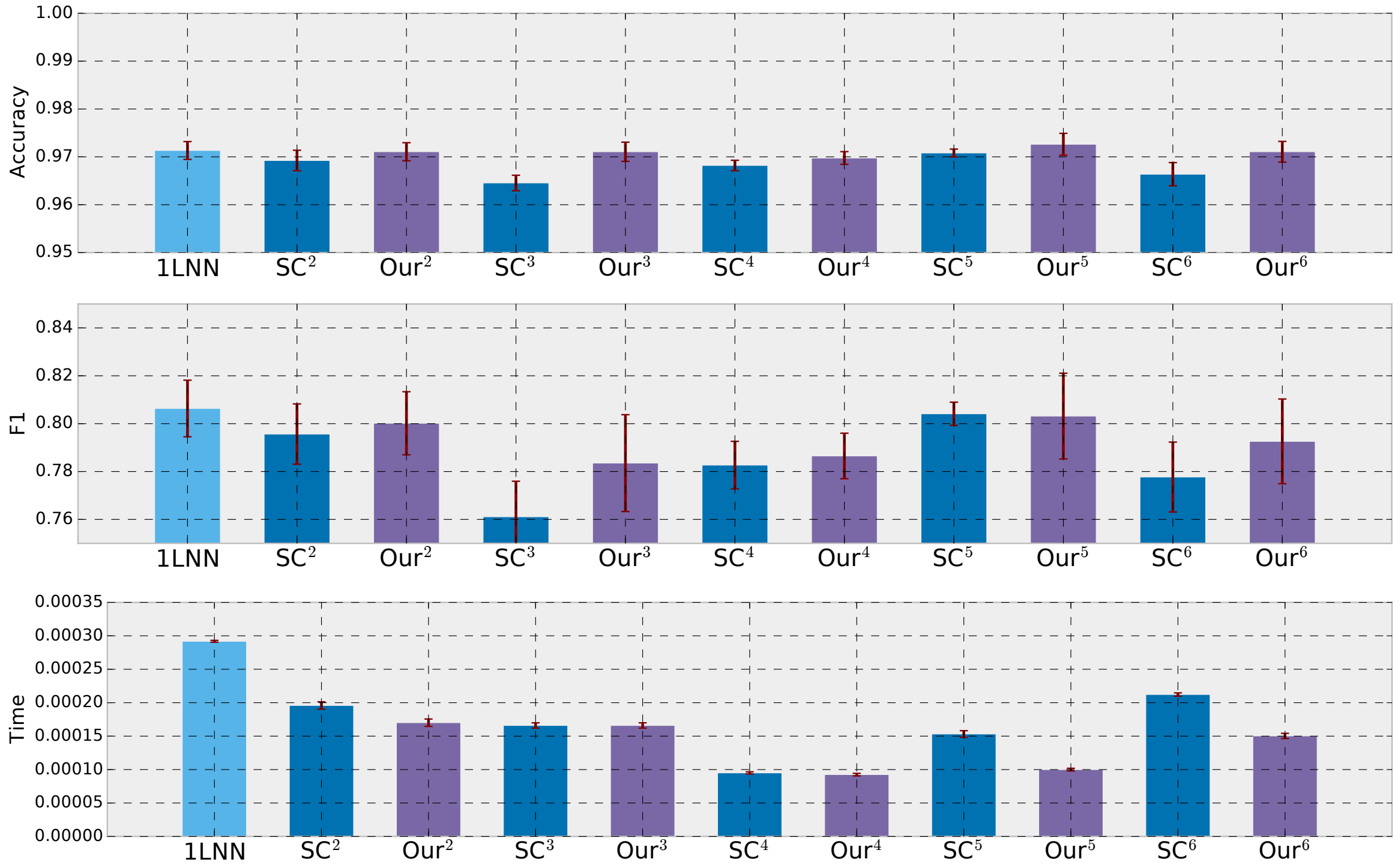

**Figure 3: Evaluation of different cascade models in terms of accuracy (top), F1 score (middle), and classification time (bottom).** `SC` **and** `Our` **correspond to the soft cascade model and our proposed firm cascade model, respectively. In all cases, the superscript** $i$ **indicates the cascade model** $Ci$**.**

We conducted these experiments using 8-fold stratified cross validation to assess generalization and computation time performance. To compare models, the cost scaling parameter $\lambda$ in our proposed firm cascade model and the soft cascade model was swept over a grid to produce a per-model speed-accuracy trade-off curve with the accuracy computed as the mean over the cross validation folds. For each model, we select the maximum accuracy point and assess the compute time required to achieve that result. The compute time that we use is the average time in seconds that each learned cascade needs to classify a test instance when operated in hard decision mode. Timing results are averaged over $10,000$ classifier evaluations [3]. We also compute the F1 score of the methods at the maximum accuracy point. All experiments were performed on 2.4GHz Intel Xeon E5-2440 CPU's.

The results are shown in Figure 3 including one-standard-error error bars. First, we can see that all of the cascaded models outperform the single-stage classifier in terms of mean classification time. When our approach is used to train the architecture $C5$, for example, we obtain a speedup of about 300% with a negligible impact on average F1 and a slight improvement in average accuracy relative to the single-stage model. We note that similar accuracy can be obtained using a single-stage two hidden layer neural network (not shown), but our model takes one quarter the time of this single-stage two hidden layer model. The soft cascade approach applied to $C5$ results in a learned model that requires about 40% more time to achieve slightly lower average accuracy and similar average F1 compared to our approach. Using a paired t-test over all folds and architectures confirms that our approach results in models that are statistically significantly faster than the soft cascade approach ($p<0.01$). A similar test applied to the accuracy results shows that our approach achieves statistically significant accuracy improvements ($p<0.01$). The F1 results are considerably more noisy and fail to achieve statistical significance despite being better on average.

In order to get a better sense of where the cost advantage of our proposed cascade model comes from compared to that of the soft cascade model framework, we take a closer look at the cascade model $C5$. We evaluate the number of cases passed through each stage (for the first cross validation fold), which in turn dictates the computation time of the cascade. The first branch of the soft cascade model passes 225 cases (out of 480 cases) through, whereas the second branch passes 282 cases through. The intersection of the two branches is a set of 136 cases, which all must run through the final stage of the classifier. On the other hand, in our proposed cascade framework, 137 and 269 points are passed through from the first and second branches, respectively. The intersection of the two branches is a set of only 86 cases, all of which must run through the final stage of the classifier. Thus, the major factor in the lower cost of our cascade framework is the lower number of cases that propagate through all stages of the cascade.

[3] Multiple runs (i.e. 10,000 classification evaluations) are performed during testing to account for the variations of the computation time

We also conducted experiments on the $C5$ model on the PuffMarker dataset (with a random train/test split of 3400/436 points) in order to evaluate the sensitivity of the results to the $\alpha$ parameter. The results in these experiments were obtained when optimizing $\lambda$ for values of $\alpha$ from the set $\{2^i : -2 \leq i \leq 10, i \in \mathbb{Z}\} \cup \{\infty\}$, where setting the value of $\alpha = \infty$ corresponds to the hard threshold gating function. As can be observed from the results shown in Figure 4, moderate values of the parameter $\alpha \in \{8, 16, 32\}$ produce optimal results in terms of both accuracy and computation time. At smaller values of $\alpha$ ($\alpha < 8$), we observe an increase in computation time. Note that at $\alpha = 0$ the gating function is not defined. On the other hand, $g_\alpha(p)$ tends to $p$ for sufficiently small (but non-zero) values of $\alpha$ (this can easily be shown via L'Hopital's rule). Hence, the performance is almost the same for sufficiently small (but non-zero) values of $\alpha$ (results for $\alpha < \frac{1}{2}$ are virtually constant). On the other hand, as we increase the value of the parameter $\alpha$ (e.g. $\alpha \geq 64$), the accuracy of the model reduces gradually. Specifically, the accuracy of the hard thresholding function ($\alpha = \infty$) is substantially lower; one hypothesis for this behavior is that the gradient of the regularizer goes to zero everywhere (except at exactly $p = 0.5$ where it is undefined) as $\alpha$ goes to $\infty$; hence, the regularizer does not contribute information to help improve the model.

As such, moderate values for the parameter $\alpha$ (e.g. between 8 and 32) produce optimal (and very similar) results for the $C5$ model on PuffMarker data set. We did not attempt to optimize the value of $\alpha$ for different cascade models as we expect this broad insensitivity to hold across models. Thus, the same value ($\alpha = 32$) has been used in all the experiments on different datasets and over different cascade models. This fixed value of $\alpha$ has been sufficient to outperform the soft cascade in all of the models over different datasets. Further optimization of $\alpha$ would only improve results further.

Finally, we note that the computation time per RMSProp iteration for the firm cascade objective is three to four times longer than for the soft cascade objective when training the same cascade architecture. However, the firm cascade objective tends to converge three to four times faster than the soft cascade objective so that the total learning time is approximately the same for both approaches.

## 4.2 Human Activity Recognition

Next, we evaluate the performance of both the proposed firm cascade model and the soft cascade model on two activity recognition (AR) datasets. The first AR dataset is the "Human Activity Recognition (HAR) using smartphones dataset" [2], where 561 features are measured from the data captured by two sensors: an accelerometer and a gyroscope, and the sensors are connected to a smartphone. The experiments have been carried out with a group of 30 volunteers, where each volunteer performed six different activities: walking, walking upstairs, walking downstairs, sitting, standing, and laying. The obtained dataset has a total of 10299 data cases. All results use 4-fold cross validation.

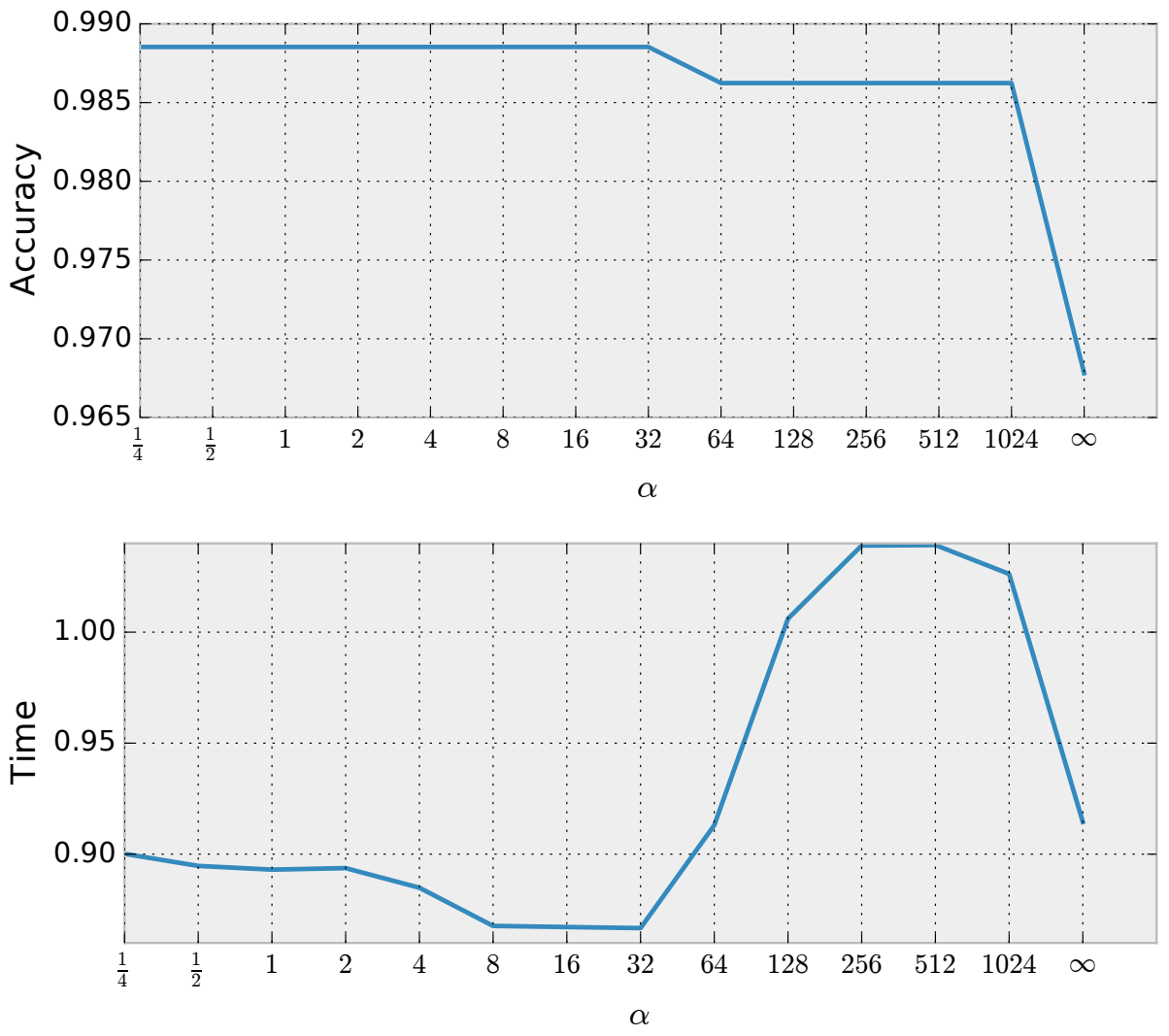

**Figure 4: Evaluation of the sensitivity of the firm cascade framework for model $C5$ versus the parameter $\alpha$: accuracy versus $\alpha$ (top), time versus $\alpha$ (bottom).**

In our experiments on cascade models, we consider walking as the positive class and the rest of the activities as the negative class, in order to produce a binary classification problem with unbalanced classes. Here we focus on the cascade models $C2$ and $C5$, and compare the results against the single-stage 1LNN (with $K = 10$), corresponding to $C1$. In $C2$, we use a LR in the first stage, and a 1LNN with $K = 10$ in the second stage. In addition, we restrict access to only the gyroscope features in the first stage, but all the features are available in the second stage. In $C5$, similar to PuffMarker experiments, we have two branches, where each branch deploys the features generated by one of the sensors (accelerometer versus gyroscope). Finally, we have LR in the branches, whereas we use 1LNN with $K = 10$ in the second stage.

The performance of the firm and soft cascade frameworks for $C1$, $C2$, and $C5$ are shown in Figure 5. In this figure, timing results are averaged over 1,000 classifier evaluations. From the results for the $C5$ model, we can observe that the values of the accuracy for both firm and soft cascade framework are similar and virtually the same as the single-stage model. However, our firm cascade framework reduces the computation time more than the soft cascade framework for both $C2$ and $C5$. The reduction in computation time is again statistically significant (p<0.01).

The second AR dataset is the "PAMAP2 physical activity monitoring dataset" [15]. PAMAP2 is recorded from 9 subjects. Subjects wore three inertial measurement units (attached to hand, ankle, and chest) and a heart rate monitor while performing 18 different activities. We consider the intensity estimation task defined in [15], where the activities are grouped according to their intensities. More specifically, we consider the vigorous activities (ascending stairs, running, and rope jumping) as the positive class, and the remaining activities (light and moderate activities) as the negative class. From the dataset and the description of the features in [15], we reproduced 119 features obtained from the three measurement units and the heart rate monitor. For our experiments on cascade models, we picked a random subset of 12,000 data cases and used 6-fold cross validation. Since in PAMAP2 we have four sensors, we use the $C5$ model with 4 branches, where each branch employs the features generated from one of the sensors. We use LR classifiers in the first-stage classifiers (i.e. classifiers in the four branches of the cascade), and a 1LNN with $K = 30$ in the second stage.

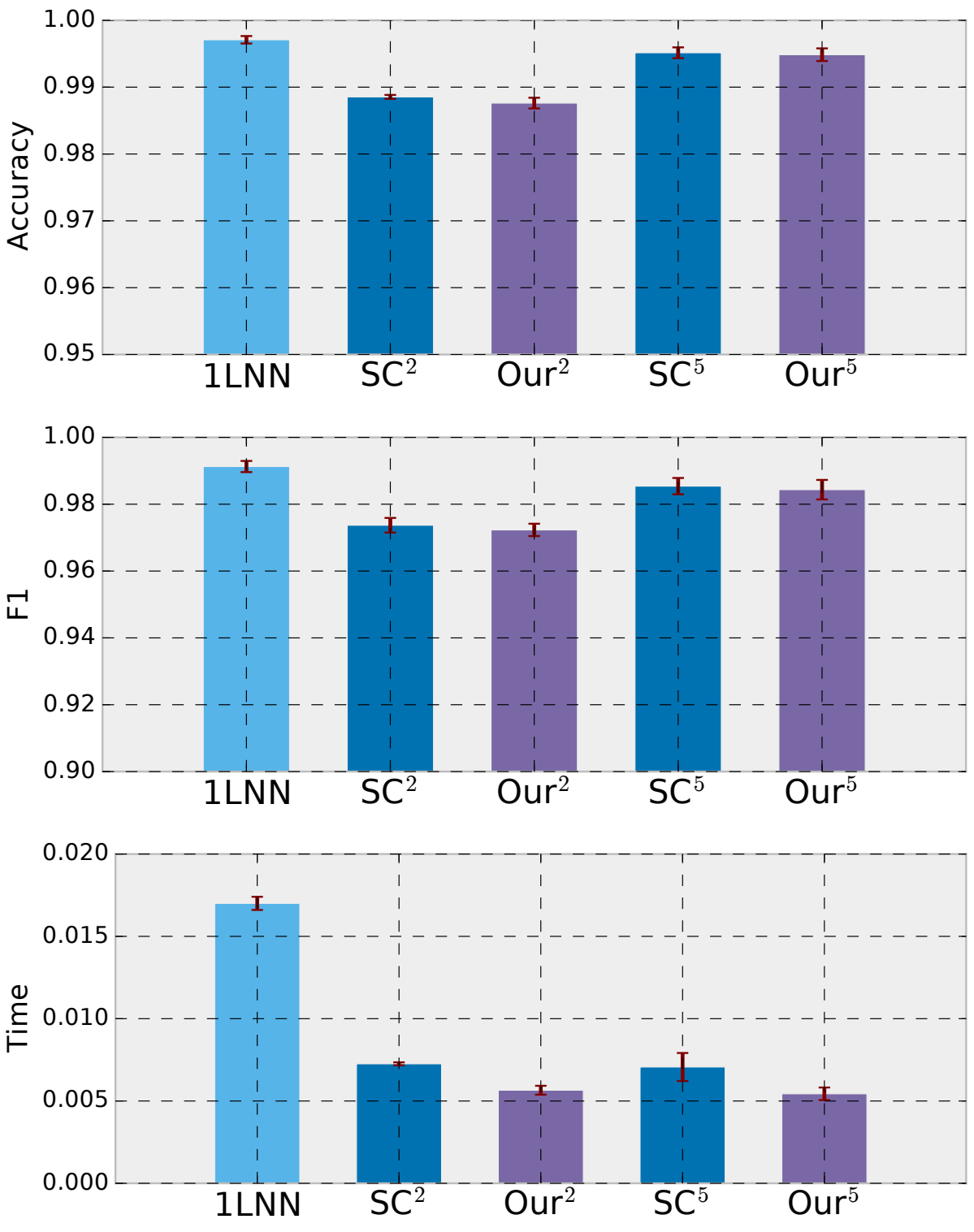

**Figure 5: Evaluation of cascade models $C1$, $C2$, and $C5$ for HAR dataset in terms of accuracy (top), F1 score (middle), and classification time (bottom). SC and Our correspond to the soft cascade model and our proposed firm cascade model, respectively. In all cases, the superscript $i$ indicates the cascade model $Ci$.**

The performance of the firm and soft cascade frameworks for $C1$ and $C5$ are shown in Figure 6. Again, timing results are averaged over 1,000 classifier evaluations. From this figure, we see that the best accuracy obtained from both firm and soft cascade frameworks are the same, but in terms of computational time, our proposed cascade framework performs substantially better. The reduction in computation time is again statistically significant (p<0.01).

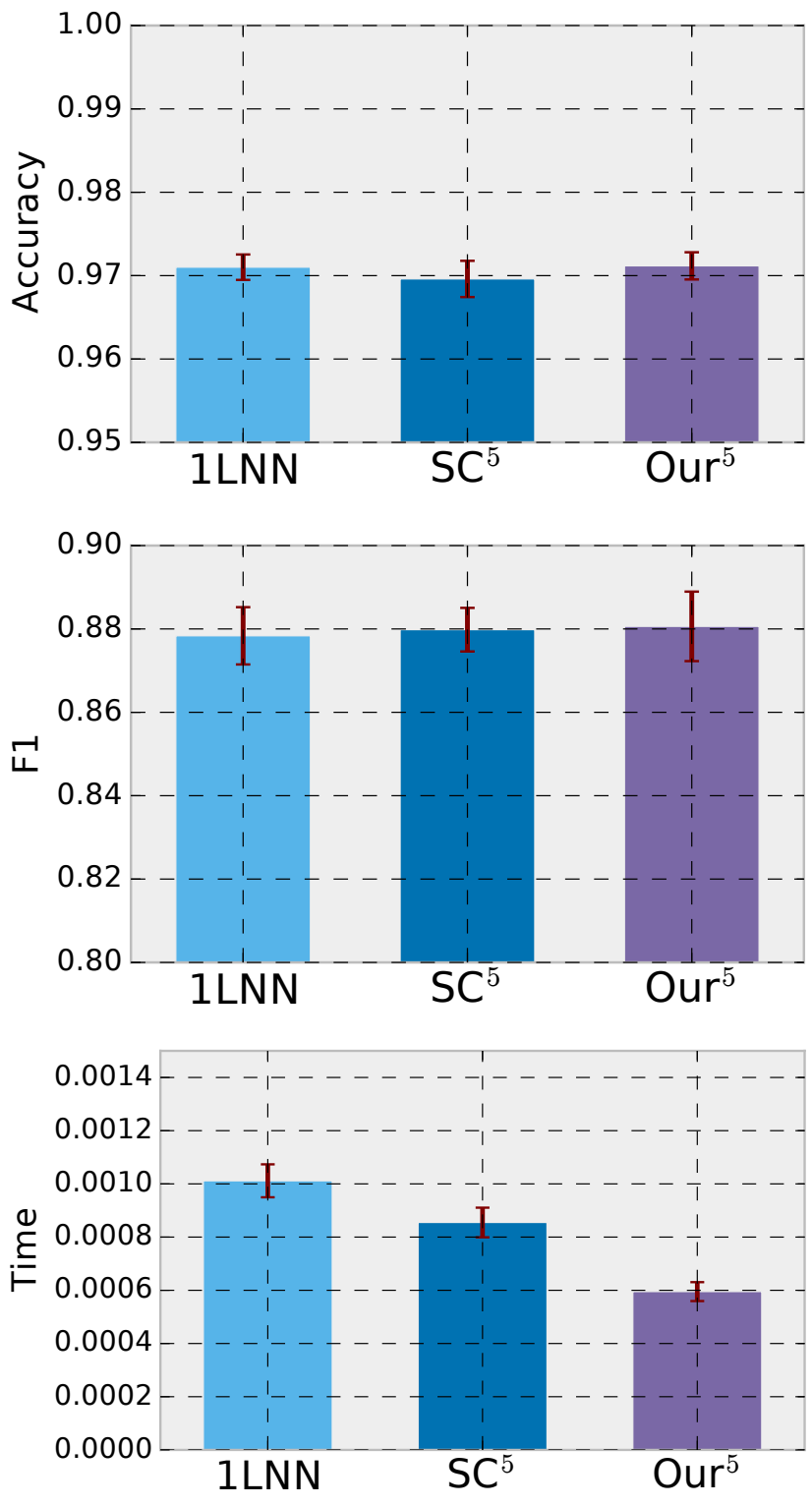

**Figure 6: Evaluation of cascade models $C1$ and $C5$ for PAMAP2 dataset in terms of accuracy (top), F1 score (middle), and classification time (bottom). SC and Our correspond to the soft cascade model and our proposed firm cascade model, respectively.**

## 5 CONCLUSIONS AND FUTURE WORK

We have introduced a new approach to cascaded classifier learning using a cascade learning objective that better matches the hard decisions that are made when the learned cascade is deployed in practice. Our results indicate that our approach can lead to significant improvements in terms of accuracy/F1-score and computational time over the soft cascade (and noisy-AND approach) on different data sets and cascade architectures. This will enabling real-time data analytics on resource-constrained networks of devices with improved accuracy and/or lower cost.

In terms of future work, we plan to develop improved cost models for the devices that we intend to deploy cascades on. The current experiments use computation time as proxy, but real applications need to consider a more general energy-based cost model that takes into consideration the cost of sensing, computing, and communicating across devices. Second, we intend to deploy the learned smoking detection models on actual hardware to assess the performance of the end-to-end system. We also plan to expand the application of the proposed architecture to other application domains and other model types. Of particular interest are more computationally intensive structured prediction-based models (for example, conditional random field models). An interesting direction is to consider adding a cloud-based stage to the architecture with much greater compute power and no resource constraints to run such models. While communicating with cloud-based computational resources over WiFi or cellular networks can be prohibitively expensive if all data must be streamed to the cloud, transmitting a small volume of cases at the end of our current cascades would be much more realistic. Finally, we note that the problem of automatically configuring a tree-structured cascade given a graph of the underlying network architecture is an interesting challenge that could further improve the speed-accuracy trade-off we have already demonstrated.

## 6 ACKNOWLEDGMENTS

The authors would like to thank Deepak Ganesan, Nazir Saleheen, and Santosh Kumar for helpful discussions of this research. This work was partially supported by the National Institutes of Health under award 1U54EB020404 and the National Science Foundation under award IIS-1350522.

## REFERENCES

[1] Amin Ahsan Ali, Syed Monowar Hossain, Karen Hovsepian, Md. Mahbubur Rahman, Kurt Plarre, and Santosh Kumar. 2012. mPuff: Automated Detection of Cigarette Smoking Puffs from Respiration Measurements. In *International Conference on Information Processing in Sensor Networks*. 269–280.

[2] Davide Anguita, Alessandro Ghio, Luca Oneto, Xavier Parra, and Jorge L. Reyes-Ortiz. 2013. A Public Domain Dataset for Human Activity Recognition Using Smartphones. In *21th European Symposium on Artificial Neural Networks, Computational Intelligence and Machine Learning, ESANN 2013*. Bruges, Belgium. https://archive.ics.uci.edu/ml/datasets/Human+Activity+Recognition+Using+Smartphones

[3] Leo Breiman. 2001. Random forests. *Machine learning* 45, 1 (2001), 5–32.

[4] M. Chen, K. Weinberger Z. Xu, O. Chapelle, and D. Kedem. 2012. Classifier Cascade for Minimizing Feature Evaluation Cost. In *International Conference on Artificial Intelligence and Statistics (AISTATS)*. La Palma, Canary Islands.

[5] Corinna Cortes and Vladimir Vapnik. 1995. Support vector networks. *Machine learning* 20, 3 (1995), 273–297.

[6] Yoav Freund and Robert E Schapire. 1997. A Decision-Theoretic Generalization of On-Line Learning and an Application to Boosting. *J. Comput. Syst. Sci.* 55, 1 (1997), 119–139.

[7] Syed Monowar Hossain, Amin Ahsan Ali, Mahbubur Rahman, Emre Ertin, David Epstein, Ashley Kennedy, Kenzie Preston, Annie Umbricht, Yixin Chen, and Santosh Kumar. 2014. Identifying drug (cocaine) intake events from acute physiological response in the presence of free-living physical activity. In *Int. Symposium on Information processing in sensor networks*. 71–82.

[8] Robert A Jacobs, Michael I Jordan, Steven J Nowlan, and Geoffrey E Hinton. 1991. Adaptive mixtures of local experts. *Neural computation* 3, 1 (1991), 79–87.

[9] Santosh Kumar, Wendy Nilsen, Misha Pavel, and Mani Srivastava. 2013. Mobile health: Revolutionizing healthcare through transdisciplinary research. *Computer* 1 (2013), 28–35.

[10] Leonidas Lefakis and François Fleuret. 2010. Joint cascade optimization using a product of boosted classifiers. In *Advances in Neural Information Processing Systems*. 1315–1323.

[11] Inbal Nahum-Shani, Shawna N Smith, Ambuj Tewari, Katie Witkiewitz, Linda M Collins, Bonnie Spring, and S Murphy. 2014. Just in time adaptive interventions (JITAIs): An organizing framework for ongoing health behavior support. *Methodology Center technical report* 14-126 (2014).

[12] Annamalai Natarajan, Abhinav Parate, Edward Gaiser, Gustavo Angarita, Robert Malison, Benjamin Marlin, and Deepak Ganesan. 2013. Detecting cocaine use with wearable electrocardiogram sensors. In *Proceedings of the ACM international joint conference on Pervasive and ubiquitous computing*. 123–132.

[13] Kurt Plarre, Andrew Raij, Syed Monowar Hossain, Amin Ahsan Ali, Motohiro Nakajima, Mustafa al'Absi, Emre Ertin, Thomas Kamarck, Santosh Kumar, Marcia Scott, and others. 2011. Continuous inference of psychological stress from sensory measurements collected in the natural environment. In *International Conference on Information Processing in Sensor Networks (IPSN)*. IEEE, 97–108.

[14] Vikas C Raykar, Balaji Krishnapuram, and Shipeng Yu. 2010. Designing efficient cascaded classifiers: tradeoff between accuracy and cost. In *ACM SIGKDD international conference on Knowledgediscovery and data mining*. ACM, 853–860.

[15] A. Reiss and D. Stricker. 2012. Creating and Benchmarking a New Dataset for Physical Activity Monitoring. In *The 5th Workshop on Affect and Behaviour Related Assistance (ABRA)*. https://archive.ics.uci.edu/ml/datasets/PAMAP2+Physical+Activity+Monitoring

[16] Mohammad Saberian and Nuno Vasconcelos. 2014. Boosting algorithms for detector cascade learning. *Journal of Machine Learning Research* 15, 1 (2014), 2569–2605.

[17] Nazir Saleheen, Amin Ahsan Ali, Syed Monowar Hossain, Hillol Sarker, Soujanya Chatterjee, Benjamin Marlin, Emre Ertin, Mustafa al'Absi, and Santosh Kumar. 2015. puffMarker: a multi-sensor approach for pinpointing the timing of first lapse in smoking cessation. In *ACM International Joint Conference on Pervasive and Ubiquitous Computing*. 999–1010.

[18] Theano Development Team. 2016. Theano: A Python framework for fast computation of mathematical expressions. *arXiv e-prints* abs/1605.02688 (May 2016). http://arxiv.org/abs/1605.02688

[19] Edison Thomaz, Irfan Essa, and Gregory D Abowd. 2015. A practical approach for recognizing eating moments with wrist-mounted inertial sensing. In *ACM International Joint Conference on Pervasive and Ubiquitous Computing*. 1029–1040.

[20] Kirill Trapeznikov and Venkatesh Saligrama. 2013. Supervised Sequential Classification Under Budget Constraints. In *International Conference on Artificial Intelligence and Statistics (AISTATS)*. 581–589.

[21] Paul Viola and Michael Jones. 2001. Rapid Object Detection using a Boosted Cascade of Simple Features. *IEEE Conference on Computer Vision and Pattern Recognition* 1 (2001), 511.

[22] Joseph Wang, Kirill Trapeznikov, and Venkatesh Saligrama. 2015. Efficient Learning by Directed Acyclic Graph For Resource Constrained Prediction. In *Advances in Neural Information Processing Systems 28*. 2152–2160.

[23] Zhixiang Xu, Matt Kusner, Minmin Chen, and Kilian Q. Weinberger. 2013. Cost-Sensitive Tree of Classifiers. In *International Conference on Machine Learning*. 133–141.

[24] Zhixiang Xu, Matt J. Kusner, Kilian Q. Weinberger, Minmin Chen, and Olivier Chapelle. 2014. Classifier Cascades and Trees for Minimizing Feature Evaluation Cost. *J. Mach. Learn. Res.* 15, 1 (2014), 2113–2144.

[25] Zhixiang Xu, Kilian Q. Weinberger, and Olivier Chapelle. 2012. The Greedy Miser: Learning under Test-time Budgets. In *International Conference on Machine Learning (ICML)*. 169.